\documentclass[journal]{IEEEtran}
\usepackage{enumitem}
\usepackage{multicol,multienum}
\usepackage{breqn}
\usepackage{stfloats}
\usepackage{multirow}
\usepackage{diagbox}
\usepackage{hyperref}
\usepackage{bm}
\usepackage{cite}
\usepackage{graphicx}
\usepackage{tabularx}
\usepackage[ruled,vlined]{algorithm2e}
\usepackage{algorithmic}
\usepackage{booktabs}
\usepackage{amsthm}

\theoremstyle{definition}

\usepackage{amssymb}
\usepackage{amsmath}
\usepackage[table]{xcolor}
\usepackage{siunitx}
\DeclareSIUnit{\mph}{mph}
\usepackage{mathtools}
\usepackage{tabularx}
\usepackage{xspace}

\usepackage{cleveref}
\crefname{section}{Section}{Sec.}
\crefname{figure}{Figure}{Fig.}
\crefname{table}{Table}{Table.}
\crefname{algorithm}{Algorithm}{Algorithm}
\crefname{equation}{Eq.}{Eq.}

\usepackage{bbding}

\usepackage[utf8]{inputenc}
\usepackage{tikz}
\usetikzlibrary{positioning}
\usetikzlibrary{shapes,arrows,arrows,positioning,fit}
\usetikzlibrary{shapes.geometric, arrows}

\usepackage[tight,footnotesize]{subfigure}
\usetikzlibrary {matrix}
\usetikzlibrary {datavisualization}

\usepackage{pifont}

\newcommand{\cparagraph}[1]{\vspace{1.5mm} \noindent \textbf{#1}\xspace}
\ifCLASSINFOpdf

\else

\fi

\newcommand\SystemName{\textsc{QuanTraffic}\xspace}

\usepackage{etoolbox}
\usepackage{pdfcomment}
\usepackage[normalem]{ulem}
\newcommand{\stkout}[1]{\ifmmode\text{\sout{\ensuremath{#1}}}\else\sout{#1}\fi}
\newcommand{\udline}[1]{\ifmmode\text{\uline{\ensuremath{#1}}}\else\uline{#1}\fi}

\newcommand{\multirowoffset}{-0.5\dimexpr \aboverulesep + \belowrulesep + \cmidrulewidth}

\begin{document}

\title{Adaptive Modeling of Uncertainties for Traffic Forecasting}

\author{Ying~Wu,
        Yongchao~Ye,
        Adnan~Zeb,
        James~J.Q.~Yu,
        Zheng~Wang
        }





\maketitle

\begin{abstract}
Deep neural networks (DNNs) have emerged as a dominant approach for developing traffic forecasting models. These models are typically trained to minimize error on averaged test cases and produce a single-point prediction, such as a scalar value for traffic speed or travel time. However, single-point predictions fail to account for prediction uncertainty that is critical for many transportation management scenarios, such as determining the best- or worst-case arrival time.
We present \SystemName, a generic framework to enhance the capability of an arbitrary DNN model for uncertainty modeling. \SystemName requires little human involvement and does not change the base DNN architecture during deployment. Instead, it automatically learns a standard quantile function during the DNN model training to produce a prediction interval for the single-point prediction. The prediction interval defines a range where the true value of the traffic prediction is likely to fall. Furthermore, \SystemName develops an adaptive scheme that dynamically adjusts the prediction interval based on the location and prediction window of the test input. We evaluated \SystemName by applying it to five representative DNN models for traffic forecasting across seven public datasets. We then compared \SystemName against five uncertainty quantification methods. Compared to the baseline uncertainty modeling techniques, \SystemName with base DNN architectures delivers consistently better and more robust performance than the existing ones on the reported datasets.

\end{abstract}

\IEEEpeerreviewmaketitle

\section{Introduction}\label{sec:intro}

Accurate prediction of future traffic information, including traffic volume, congestion levels, traffic speed, and travel time, is crucial for a variety of transportation applications, such as congestion control \cite{akhtar2021review}, travel time estimation\cite{9905416}, emergent route planning\cite{tian2018research}, and taxi demand prediction\cite{9945629}. It also supports effective transportation planning and management, enabling policy-makers to optimize traffic flow and emergency response planning while providing road users with a safer and more efficient travel experience.

In recent years, there has been a growing interest in the development of advanced predictive models for traffic information using data-driven approaches such as deep neural networks (\textbf{DNNs}) \cite{9112608}.  Multiple studies have shown that traffic forecasting models based on DNNs outperform classical machine learning methods by a large margin \cite{9352246, DL2021Jiang}. However, these models are typically trained to minimize the averaged prediction error, resulting in a considerable variation in performance across test samples \cite{SurveyJohnson}. This variation poses significant challenges for individual use cases where precise traffic prediction is critical but difficult to achieve. For instance, accurately predicting traffics at crossroads with high traffic volumes during rush hours is crucial for urban travel but can be challenging due to complex traffic patterns \cite{thulasidasan2019mixup}. As the accuracy of traffic forecasting can fluctuate over time \cite{3403294}, it is crucial to model and quantify the prediction uncertainty of the model for individual roads, locations, and sensors for a given prediction window (e.g., traffic predictions for the next $n$ minutes).

Modeling and quantifying traffic forecasting uncertainties have real-world use cases in transportation management. For example, in emergency response situations, having knowledge of the upper bound of a traffic flow prediction can help emergency responders avoid congested roads and take the quickest and safest route to their destination. Similarly, knowing the earliest and worst-case travel times enables users to make informed decisions about their travel plans, such as selecting the most convenient transportation method while minimizing the likelihood of being late to an appointment or arriving too early. Therefore, by quantifying the forecasting uncertainty, we can improve the reliability of traffic prediction and traffic management efficiency. Unfortunately, despite the huge benefits of uncertainty modeling, prior research on traffic forecasting has largely overlooked this issue. This is a massively missed opportunity.

The recent studies presented in \cite{Estimating2021} and \cite{Qian2023} were among the first attempt to model uncertainties in traffic forecasting. The work presented in \cite{Estimating2021} uses a classical statistical method to quantify the uncertainties associated with average daily traffic volume forecasts. However, this approach requires manual tuning and selecting a set of features for each dataset to fit a linear model. Its requirement of intensive expert involvement thus limits the practicability. 
In \cite{Qian2023}, a method based on variational inference and deep ensembling  is employed to estimate both the data and model uncertainties in traffic forecasting. This approach provides a more comprehensive understanding of the uncertainties involved, but it requires significant changes to the original model architectures, which limits its generalization ability.

In this paper, we present \SystemName, a generic framework for quantifying the prediction uncertainties of a DNN-based traffic forecasting model.  Unlike prior work \cite{Estimating2021,Qian2023}, \SystemName is designed to minimize engineering efforts and expert involvement. It  can work with any DNN model without changing the underlying architecture during deployment. By producing a prediction interval 
{(\textbf{PI})} that captures the range in which the true value (such as travel time) is likely to fall, \SystemName enhances the capability of a standard DNN to capture prediction uncertainties. 

At the core of \SystemName is a quantile function built upon the recently proposed Conformalized Quantile Regression (\textbf{CQR})  algorithm~\cite{romano2019conformalized}. The quantile function estimates the PI of a given model output based on the data distribution and validation errors observed during the standard DNN model training process. \SystemName is designed to simplify the training and usage of the quantile function. The process involves attaching a linear layer to the last layer of the base DNN model and using a pinball loss function \cite{Koenker2001Quantile} during standard DNN training. Once trained, the base DNN model and the quantile function can be used as standalone components during deployment. During inference, the quantile function generates an initial PI based on the DNN model's single-point prediction. Then, an adjustment is made to refine it and improve its accuracy. The goal is to increase the coverage of the PI while simultaneously minimizing the width between its upper and lower bounds.

Unlike standard CQR that uses a global constant value to adjust the initial PI, \SystemName develops an adaptive scheme to tailor the adjustment value applied to the initial PI for the specific location (or sensor node) of the test sample within a given prediction window. This allows the uncertainty method to consider the prediction difficulty of each test sample. For example, locations that are known to have high variability in traffic patterns may require a larger adjustment to achieve accurate PIs, while those with more predictable traffic patterns may require a smaller  value. We achieve this by utilizing a calibration table that is automatically constructed using an optimization function on a calibration dataset. This table provides the optimal adjustment for a node-prediction-window combination, enabling \SystemName to account for unique prediction challenges in each test sample. By using differentiated residues to adjust the initial PI, we achieve greater precision and reliability over the standard CQR. 

We have implemented a working prototype of \SystemName, which will be open-sourced upon acceptance of this work. We evaluate \SystemName by applying it to five representative DNN architectures \cite{Yu2018spatio, ijcai_WuPLJZ19, aaai_ZhengFW020,sigir_LaiCYL18,  wu2020connect} for traffic forecasting. We then test the \SystemName-enhanced DNN model on seven public datasets for traffic speed and flow prediction. We compare \SystemName against five state-of-the-art uncertainty modeling methods \cite{Vladimir2005,Blundell2015weight, pmlr-v48-gal16, Roger1978, Mach2008Tutorial} and a classical method based on historical data. Experimental results show that \SystemName consistently outperforms competing baselines across DNN models and datasets, delivering better and more robust performance for uncertainty quantification. 

This paper makes the following contributions:
\begin{itemize}
\item It presents a generic framework to model prediction uncertainty for DNN-based traffic forecasting models, requiring no change to the base DNN model architecture (\cref{sec:method}).
\item It develops an adaptive scheme to tackle the prediction challenge of individual locations, leading to more robust results than standard uncertainty modeling methods (\cref{subsec: component}).
\item It provides a large independent study to highlight the importance of uncertainty modeling of traffic forecasting. We hope our study can encourage further research along this line (\cref{sec:results}). 
\end{itemize}

\vspace{-4mm}
\section{Motivation\label{sec:motivation}}
\vspace{-2mm}
\begin{figure}
    \centering
    \includegraphics[width=0.98\linewidth]{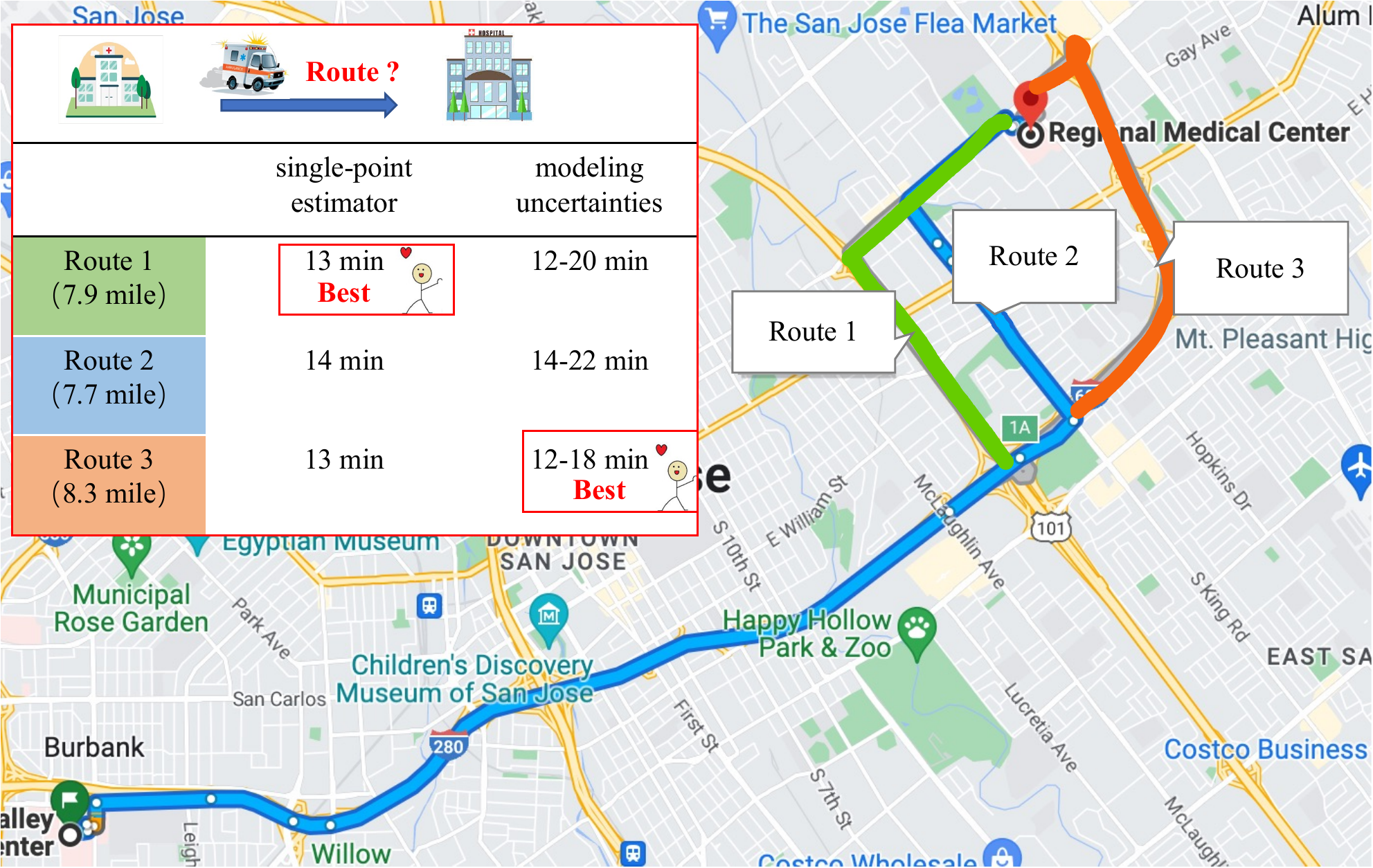}
    \caption{\textbf{A representative emergency route planning example between two points.} A single-point travel time estimation does not provide the upper-bound travel time for route selection which is important for ensuring the worst-case arrival time.}
    \label{fig:motivation}
\end{figure}

As a motivative example, consider the emergency route planning task depicted in \cref{fig:motivation}. This example represents a day-to-day scenario where traffic forecasts are employed to plan optimal routes, like an ambulance traveling from Regional Medical Center to Santa Clara Valley Medical Center. The objective is to identify the quickest route based on the predicted traffic while ensuring the worst-case arrival time.

Typical traffic forecasting models can provide estimates for travel time of different routes, as seen in \cref{fig:motivation}. However, relying solely on these single-point forecasts for route planning can lead to unreliable decisions since they lack information about the uncertainty or confidence of the predictions. In the example shown, a user may choose Route 1 due to its lowest average travel time (13 minutes). Nevertheless, this decision overlooks the potential variability in traffic dynamics and route complexity, possibly rendering a worse real travel time. To account for this variability and provide worst-case scenario estimates, it is essential to have confidence intervals associated with the travel time predictions. Incorporating such confidence estimation can have a significant impact on time-critical route planning tasks, which is overlooked in the current literature.

We show how estimating the uncertainty of travel time predictions can provide valuable information for time-critical route planning. By providing travel time bounds besides single-point forecasts, we can tell that selected routes have much higher variability in their predicted travel times than others. If the uncertainty estimation is accurate, the additional information help users avoid routes that are more uncertain on travel times, even if their average predicted travel time is lower. For example, in the scenario depicted in \cref{fig:motivation}, one may finally choose Route 3 despite it being the longest, in order to ensure the worst-case arrival time.

This example demonstrates the limitations of single-point-based traffic forecasting models for time-critical travel planning. This highlights the need for uncertainty modeling in traffic forecasting. To address this, our work proposes a generic framework for precise single-point and bound estimations to better model traffic prediction uncertainties.

\section{Background and Related Work}\label{sec:related}

Our work builds upon the following past foundations, but our focus differs from each.

\subsection{Spatio-Temporal Traffic Forecasting}\label{subsec:st traffic forecasting}

Traffic forecasting is a well-established research topic with a wide range of proposed solutions. Classical statistical methods, such as historical average, regression, and integrated moving average models, have been explored in the past \cite{Box1970}. However, more recent research has leveraged DNNs to model the spatio-temporal correlations in traffic data. Compared to classical statistical methods, DNNs can better capture complex relationships in historical data while avoiding the need for hand-engineered features. Researchers have explored several approaches to represent traffic data, including temporal sequence modeling with recurrent neural networks~\cite{8614060}, multidimensional matrix representations with convolutional neural networks \cite{8526506}, and graph neural networks \cite{jiang2021dl}. DNN-based methods have been shown to deliver state-of-the-art results in various traffic-related tasks, such as ride-sharing \cite{9254149}, and travel planning \cite{CHEN2022126060}. Due to the better performance over alternative methods, DNNs have emerged as the dominant approach for building traffic forecasting models.

The majority of existing DNN-based traffic forecasting models only provide a single-point prediction such as the travel time. However, a single-point estimation only reflects the average traffic scenario but not the best or worse cases.
As highlighted in Section \ref{sec:motivation}, the upper and lower bounds of the travel time can be critical for choosing the best route in time-critical route planning tasks. This requires one to consider the uncertainty of the predicted travel time and to produce metrics similar to statistical confidence intervals. This work aims to address this issue by developing a generic approach that can provide such information from any DNN model, making it applicable to a wide range of traffic forecasting architectures.

\subsection{Modeling Prediction Uncertainty}\label{subsec:uncertainty}
Although uncertainty modeling has been largely overlooked in prior traffic forecasting approaches, it has drawn much attention in other DNN-based modeling tasks. Various techniques have been proposed to quantify prediction errors, confidences, or uncertainties. These methods can be broadly categorized as Bayesian and Frequentist ones, which have been extensively studied in the literature \cite{arXiv2021_03342, pmlr-v80-pearce18a}.

\cparagraph{Bayesian models} provide a robust probabilistic framework for modeling uncertainty with Bayesian statistics \cite{journals/neco/MacKay92a}. In this approach, the model incorporates prior knowledge or beliefs for parameter initialization and infers the posterior distribution using the likelihood function between the data and a predefined initial distribution. Techniques for quantifying uncertainty in DNNs using Bayesian models include Monte Carlo (MC) dropout \cite{pmlr-v48-gal16} and Variational Inference \cite{ Blundell2015weight}. However, there are emerging challenges, such as relatively low computation efficiency and strong prior distribution assumptions. These challenges can be particularly acute in high-dimensional models or large datasets.

\cparagraph{Frequentist methods} provide predictions based on a single forward pass with a deterministic network and quantify uncertainty by using additional qualification schemes. These methods use post-hoc calibrations, such as conformal prediction and differentiable modeling structures, and their loss objectives, such as quantile prediction \cite{nips_ChungNCS21}, to capture uncertainties. Ensemble methods, such as those that use random initialization or a mixture of experts \cite{nips_Lakshminarayanan17}, retrain models on partial datasets, or adopt data augmentation techniques (e.g., cross-validation \cite{iclr_RitterBB18}, and bootstrap aggregating \cite{aaai_HuangZH19}), are also considered frequentist methods. However, ensemble methods require trial-and-error adjustments to parameters without a solid mathematical foundation, leading to poor coverage guarantee. Moreover, frequentist methods are often overconfident \cite{nips_ChungNCS21, stankeviciute2021conformal}, which can result in inaccurate uncertainty estimates. Despite these limitations, frequentist methods are attractive because they are computationally efficient and do not require prior assumptions on the model or data distribution.

\subsection{Summary}

In summary, existing methods for quantifying uncertainty suffer from several issues, including inaccurate coverage guarantees, strong distributional assumptions, and insufficiently calibrated prediction intervals. These challenges are further compounded when dealing with heteroscedastic data\footnote{Heteroscedastic data refers to data that has varying levels of variability or scatter across its range, as opposed to homoscedastic data which has consistent levels of variability or scatter. An example of heteroscedastic data in traffic forecasting is rush hour traffic, which typically has more variability than traffic during off-peak times.}.

To address these issues, we propose an adaptive conformalized quantile model that provides a unified and reliable framework for quantifying uncertainty in traffic forecasting. Our approach is one of the first attempts to use frequentist methods for estimating uncertainty in traffic forecasting. We provide a comprehensive and structured comparison of existing approaches on real traffic data, using a variety of state-of-the-art DNN-based traffic forecasting models.
We hope our work  can promote more research in this important area of uncertainty quantification for traffic forecasting. 
\section{Preliminaries}\label{sec:problem}

\subsection{Problem Definition}\label{sec:problem/define}

A single-point traffic forecasting model attempts to predict future traffic information based on the past. Examples of traffic information include flow, speed, and density. Given a $D$-dimensional multivariate time series $X = \begin{Bmatrix} x_{t-(m-1)},\hdots, x_{t-1} ,x_{t} \end{Bmatrix} \in \mathbb{R}^{ N \times D}$ collected at $t$ time with past $m$ steps from $N$ data sources (roads or senors), a point forecasting model $f(\cdot)$ attempts to estimate the multivariate time series $Y = \begin{Bmatrix} y_{t}, \hdots, y_{t+h-1}, y_{t+h} \end{Bmatrix} \in \mathbb{R}^{ N \times D}$ in the next prediciton window (e.g., $h$ steps), : $X \overset{f(\cdot)}{\rightarrow} Y$.

Probabilistic traffic forecasting involves predicting the likelihood of various potential outcomes, rather than simply estimating the most likely outcome. Given historical data $X$, a probability predictor $F(\cdot)$ can estimate the uncertainty of future traffic conditions by producing a set of PIs, which contain the real data with a certain level of confidence.

Specifically, a probabilistic forecasting model attempts to estimate PIs for a set of future values denoted as $\hat{Y} = \{\hat{y}_{t}, \ldots, \hat{y}_{t+h}\} \in \mathbb{R}^{2 \times N \times D}$, where $t$ is the current time, $h$ is the prediction window. The PI for $\hat{y}_{t+h}$ is defined as ${\hat{y}_{t+h}^{l} \leq \hat{y}_{t+h} \leq \hat{y}_{t+h}^{u}}$, where $\hat{y}_{t+h}^{l}$ and $\hat{y}_{t+h}^{u}$ are the lower and upper bounds of the PI, respectively. By specifying a confidence level $\alpha$, we can say that there is a probability of $(1-\alpha)$  that the true value $Y_{t+h}$ falls within the PI $\hat{Y}_{t+h}$. The probabilistic forecasting process can be described as $X \overset{F}{\rightarrow} \hat{Y}$, where $F$ is the probability predictor that maps historical data $X$ to the predicted PIs for $\hat{Y}$.

This paper aims to develop a generic framework to extend a single-point predictor $f$ to model the
uncertainties in traffic modeling tasks.  We target DNN-based traffic forecasting models as they are the core of
state-of-the-art traffic forecasting methods~\cite{jiang2021dl}. It is worth nothing that our approach does not change
the underlying DNN model structure and can be integrated with other classical supervised learning methods, including
support vector machines~\cite{708428}, linear~\cite{PAPADOPOULOS2011842} and non-linear regressors~\cite{Random2001}.

\subsection{Optimization Goals\label{sec:goal}}
Our goal is to generate PIs that are both narrow and accurate, meaning that they are just wide enough to cover the true value (e.g., the actual traffic speed) with high probability. To achieve this, we need the PI to be \emph{discriminative} so that it is narrow for single-point predictions with high confidence, but wide enough to ensure good coverage overall \cite{NIPS2021_00225}.

We also require the uncertainty method to perform well across test samples and have good coverage, which means that we need the results to be \emph{valid}~\cite{NIPS2021_00225}. Specifically, for a given probability or coverage rate of $x$ (defined as $1 - \alpha$), we expect the PI to encompass the true value at least $x$ of the time.

\subsection{Quantified Metrics\label{sec:qm}}

We adopt the established practices of prior uncertainty modeling methods in related domains~\cite{NIPS2021_00225,pmlr-v80-pearce18a} and use two metrics to evaluate the efficacy of the PI: \textbf{validity} and \textbf{discrimination}~\cite{NIPS2021_00225}. Specifically, we use \emph{mean prediction interval width (MPIW)} and \emph{prediction interval coverage probability (PICP)} to measure these requirements. Our goal is to obtain a small MPIW with adequate coverage across test samples. While a large MPIW can ensure good coverage, it may frequently underestimate or overestimate traffic information, even when the model is relatively confident with the prediction. On the other hand, a small MPIW can miss the true value, leading to incorrect decisions such as selecting a slower route due to incorrect predictions. 

\cparagraph{Mean prediction interval width} We compute the MPIW across $n$ test samples as:
\begin{equation}
    \label{eq:PI}
        \mathrm{MPIW} = \frac{1}{n} \sum_{i=1}^{n} \begin{vmatrix}y_i^{u} - y_i^{l}\end{vmatrix}.
    \end{equation}
The MPIW is a \emph{smaller-is-better} metric as we anticipate the prediction width to be as narrow as possible.

\cparagraph{Prediction interval coverage probability} In addition to MPIW, we also consider the frequency PI covers the real value over the complete $n$ test samples. The observed coverage $\hat{C}_{\alpha}$ is computed as:
\begin{equation}
    \label{eq:coverage}
        \begin{matrix}
            \hat{C}_{\alpha} = \frac{1}{n}\sum_{i=1}^{n}c_{i},
    &
    c_{i} = \left\{\begin{matrix}
     1, & y_{i} \in [y_i^{l},y_i^{u}]\\
        0, & y_{i} \notin  [y_i^{l},y_i^{u}]
        \end{matrix}. \right.
    \end{matrix}
    \end{equation}
Here, $c_i$ is a binary value counting if the true value $y_i$, is within the PI $\{y_i^{l},y_i^{u}\}$ for test sample $i$.

Given a target probability level $1 - \alpha$, the \emph{expected coverage} is $1 - \alpha$ for $n$ test samples. $\alpha$ is defined as mis-coverage rate. For this case, an ideal uncertainty predictor produces $n$ PIs for all test samples to cover the real value at least $1-\alpha$
time, rendering the \emph{observed coverage} $\hat{C}_{\alpha}$, to be $1 - \alpha$ as well.
If $\hat{C}_{\alpha}$ is less than $1 - \alpha$, the PI is likely undercovered.

Undercoverage is generally bad since it limits the PI to meet the requirement of retaining the true values of important test samples.

\subsection{Conformalized Quantile Regression}
\label{subsec: CQR}

Our approach is inspired by the recently proposed CQR
algorithm~\cite{romano2019conformalized}. CQR combines conformal prediction \cite{ Vladimir2005}
and classical quantile regression for uncertainty qualification. It uses quantile regression to generate an initial PI
and then uses conformal prediction to adjust the PI if required.

\subsubsection{Working mechanism of CQR}
In layperson's terms, a quantile regressor is a tool for quantifying the chance of having a specific point in a range of possible outcomes. For instance, if we want to predict the 90th percentile or quantile for traffic speed, it means there's a 5\% chance that the actual speed will be lower than our prediction, and an 5\% chance it will be higher. By fitting two quantile functions - one at the 5th percentile and another at the 95th percentile - we can create a prediction interval (PI) with a 90\% coverage. This is where CQR comes in handy for modeling traffic uncertainty, as it allows us to estimate these quantiles. However, no previous work has used CQR for traffic uncertainty modeling, making our work the first attempt to fill this research gap.

In CQR, the training data is divided into two distinct sets: a training set and a calibration set. The training set is used to develop the traffic forecasting model using standard supervised learning techniques. Then, two quantile regressors are trained on the training set to generate initial estimates of the upper and lower bounds of the PI. One quantile regressor is used for estimating the upper bound quantile, and the other for the lower bound. After obtaining the initial estimates, a conformal step is performed on the calibration to learn a \emph{single} adjustment (e.g., increasing or decreasing the PI by x\%) to be used to adjust the PI for all test samples. 

\subsubsection{Limitations of CQR\label{sec:lmcqr}} 
A drawback of using CQR for traffic forecasting is that it only produces a single, global quantile adjustment for all test samples. However, a global quantile adjustment is unlikely to be effective, as
prediction difficulties can vary significantly across domains
or roads. To illustrate this point, consider once again Routes 1 and 2 given in our motivation example (\cref{sec:motivation}). Suppose Route 1 is more complex than Route 2, which is smoother. Applying a global adjustment, such as increasing or decreasing the model prediction by a certain percentage, to both Routes 1 and 2 is unlikely to be effective. It could result in a PI that is too narrow for the complex Route 1 or too wide for the smooth Route 2.  A better approach, and is the one adopted in this paper, is to generate different adjustment ranges for each route, resulting in more accurate and reliable PIs. Doing so  enhances the robustness of traffic uncertainty modeling by considering the specific prediction difficulties of each location.

\section{Our Approach}
\label{sec:method}

\begin{figure*}
    \centering
    \includegraphics[width=0.80\textwidth]{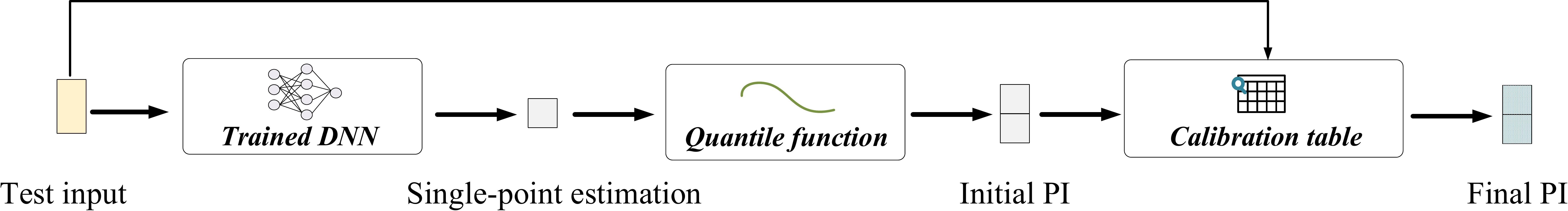}
    \caption{\textbf{Modeling forecasting uncertainties during deployment.} \SystemName extends the standard DNN-based traffic forecasting model to quantify prediction uncertainties. For a given a test input, a trained DNN generates a single-point prediction akin to traditional DNN models. This single-point prediction is subsequently processed by a quantile function, which produces an initial prediction interval (PI) for the input. The initial PI, along with the test input, is then passed through the calibration component, to produce a final PI. }
    \label{fig:deployment}
\end{figure*}

\begin{figure*}
    \centering
    \includegraphics[width=\textwidth]{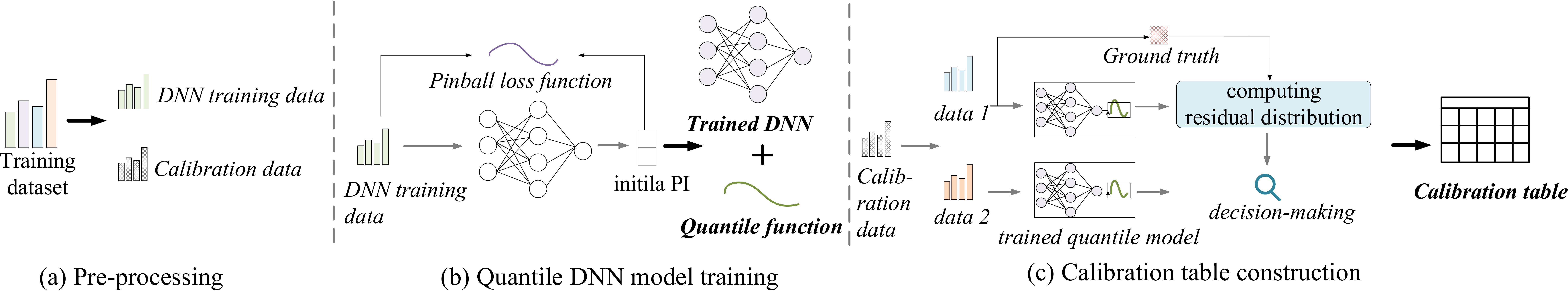}
    \caption{\textbf{Training workflow of our approach.}
    We first partition the DNN model training datasets into disjoint parts (a). We use the first part to train the DNN forecasting model and a quantile function that can produce an initial PI during deployment (b). We then apply the trained DNN model and quantile function to the set-aside calibration dataset to build the calibration component (c). }
\label{fig:training}
\end{figure*}

\SystemName is a generic framework for quantifying the prediction uncertainties of DNN-based traffic forecasting models.  Given a certain level of confidence, the \SystemName framework computes a PI that defines a range of possible values where the real data (e.g., traffic time, flow or speed) is likely to fall within.

\SystemName extends upon the existing CQR framework but provides a significant advantage by using a dedicated calibration component to dynamically adjust the width of the PI  based on locations and prediction windows (e.g., traffic conditions in the next $n$ minutes, also known as \emph{prediction horizon}), leading to improved accuracy. Another important feature of \SystemName is that it does not require any modifications to the underlying DNN model structure, making it  easy to implement. Instead, \SystemName uses a dedicated loss function to train the DNN model and a quantile function to predict an initial PI. This allows \SystemName to be seamlessly integrated with any DNN-based traffic predictor, providing an efficient and effective solution for improved traffic forecasting.

This section begins by providing an overview of \SystemName and illustrating its design principle. It then discusses how to train and construct the key components that drive the DNN model to achieve better coverage and more accurate PIs and finally elaborates on how  uncertainty qualification is achieved with test samples during deployment.

\subsection{Overview of \SystemName}
Figure \ref{fig:deployment} demonstrates how \SystemName enhances a standard DNN-based traffic forecasting model to quantify uncertainties for traffic forecasting during deployment. For a test input, the trained DNN model first produces a single-point estimation for, e.g., speeds or travel time, just like any existing traffic forecasting model. This single-point prediction is then passed to a quantile function to produce an initial PI. This PI is a 2-dimensional numerical vector, $\{l, u\}$, where $l$ is the lower bound and $u$ is the upper bound value.

In contrast to standard CQR that applies a global adjustment to the initial PI, our approach utilizes the test input's sensor node id and prediction window to locate a residual value, $\delta_i$, from a calibration table. This value is specific to the given node id, $i$, and the target prediction window, which is likely to provide a good trade-off between coverage and discrimination (see Section \ref{sec:qm}). After retrieving $\delta_i$, we then apply it to adjust the width of the initial PI, as $\{l-\delta_i, u+\delta_i\}$, to obtain the final PI. Through this process, the PI can be more accurately adjusted to reflect the specific context of the prediction, leading to enhanced accuracy and better coverage.

As shown in Figure \ref {fig:training}, our approach adapts the standard DNN model training workflow for uncertainty modeling. Our training process starts by partitioning the training data into two \emph{disjoint}, namely \emph{training} and \emph{calibration} sets respectively, to train the DNN model and build the calibration table separately. This process is described in Section \ref{subsec:Pre-processing}. Specifically, the first traing set is used to train both the DNN-based traffic forecasting model and a quantile function, using a pinball function to compute the loss during training (Section \ref{subsec:Quantile}). Once the DNN model and quantile function have been trained, we then move on to use the calibration set to construct the calibration table (Section \ref{subsec: component}).

\subsection{Pre-processing of Training Data}\label{subsec:Pre-processing}
As shown in \cref{fig:training} (a), as a \emph{one-off} preprocessing step, the training data is split into a standard model training dataset and a calibration dataset.
Specifically, in this paper, we set aside 40\% of the training data for calibration purposes, but this ratio can be adjusted by the user. The calibration dataset is created using random sampling on nodes, ensuring that it contains data for each node (or sensor), if possible. During the preprocessing stage, the quantile function (e.g., a linear function) is also added to the last layer of the DNN model.

\subsection{Training DNN Model and Quantile Function\label{subsec:Quantile}}

As depicted in \cref{fig:training} (b), the underlying DNN model and the attached quantile function are trained as a single network using the DNN training data. Then, during the back-propagation, the pinball loss is used to figure out the loss of the network. Specifically, the native DNN model makes a single-point estimate for each input, which is then passed to the quantile layer to determine the upper and lower boundaries of the prediction interval. Then the loss of the network is computed as follows:

\begin{equation}
	\label{eq: pinball loss}
    L_{\alpha} =\left\{\begin{matrix}
\alpha_{\textnormal{tra}}(y-\hat y) & y-\hat y> 0 \\
(1 - \alpha_{\textnormal{tra}})(\hat y - y) & \textup{otherwise}
\end{matrix}\right.,
\end{equation}
where $\alpha_{\textnormal{tra}}$ is a pre-defined mis-coverage rate (see Section \ref {sec:qm}) used at the training phase, $y$ is the ground truth, and $\hat{y}$ is the single-point prediction given by the base DNN model.

Essentially, the pinball loss function applies different weights to positive and negative residuals based on a known level of confidence ($1-\alpha$) \cite{Koenker2001Quantile}. A smaller value of $\alpha$ results in greater penalty for samples with values smaller than the predicted value, and vice versa \cite{RePEc:spr:aodasc:v:9:y:2022:i:2:d:10.1007_s40745-020-00253-5}. When $\alpha_{\textnormal{tra}}$ is set to 0.5, \cref{eq: pinball loss} degenerates to the yielding the median value of the corresponding dataset. Practically, setting $\alpha_{\textnormal{tra}}$ to $0.1$ expects that $90\%$ of ground truths would fall within the generated PI.

\subsection{Building Calibration Table}\label{subsec: component}

In Figure \ref{fig:training} (c), we outline the process of constructing the calibration table using the left-over calibration dataset.

\subsubsection{Calibration table} The proposed calibration table is a matrix designed to record the residual errors of a sensor node (or location) for a given prediction window. Each row of the table stores the residuals of a specific sensor node for various prediction windows, while each column records the residuals of a particular prediction window for different sensor nodes.

\subsubsection{Data preprocessing} Similar to the standard DNN training procedure, where the training data is partitioned into training and validation sets, we split the calibration data into two sets: $\chi_{1}$ for constructing the potential adjustments set for the initial PI and $\chi_{2}$ for decision-making to determine the best adjustment. In the first set $\chi_{1}$, the trained quantile DNN predicts the traffic information for each test node in a given prediction window. Then the prediction errors or residuals is calculated by measuring how far the PI boundaries deviate from the ground truth. In the second validation set $\chi_{2}$, the residual candidates obtained in $\chi_{1}$ are used to determine the optimal values. This optimal values is stored in the corresponding cells of the calibration table for a given pair of node id and prediction window. We gain this optimal values by minimizing a carefully designed objective function. In this paper, we leave 50\% of the calibration data as the $\chi_{2}$.

\subsubsection{Compute residual percentiles}
To compute the residuals for populating the calibration table, we apply the trained quantile DNN to each test sample (from the first calibration dataset $\chi_{1}$) of a given node $i$ and prediction window to produce a PI, denoted as $\{\hat{y}_{i,j}^{l},\hat{y}_{i,j}^{u}\}$, where $\hat{y}_{i,j}^{l}$ and $\hat{y}_{i,j}^{u}$ are the lower and the upper bound of
the PI of node $i$ in the prediction window $j$,  respectively. We then compute the residuals of the PI as $R_{i,j} = \left| \hat{y}_{i,j}^{u} - y_{i,j} \right|\cap \left| \hat{y}_{i,j}^{l} - y_{i,j} \right|$, where $y_{i,j}$ is the true value of node $i$ in the given prediction
window $j$. Note that there may be multiple test samples for a given node-prediction window pair, resulting in $n$ residuals denoted as $R^1_{i,j}, R^2_{i,j}, \dots, R^n_{i,j}$. 

Next, the percentiles of $n$ residuals for each node-prediction-window combination could be constructed. Specifically, the residual candidates $R^1_{i,j}, R^2_{i,j}, \dots, R^n_{i,j}$ are first sorted by value and subsequently partitioned into $m$ equal-size groups, commonly referred to as percentiles, based on their relative position within the distribution. It is worth noting that the number of quantiles $m$ is a user-defined parameter that balances calibration accuracy and computational complexity. For our study, we set $m=100$. The 0th percentile corresponds to the lowest value of residuals, while the 100th percentile corresponds to the maximum value in the residual datasetCompared to histogram-based approaches, residual quantiles demonstrate more robustness to outliers (as demonstrated in \cref{sub:gvq}). Finally the residual percentiles are denoted as $Q^1_{i,j}, Q^2_{i,j}, \dots, Q^m_{i,j}$.

\subsubsection{Choosing the best quantile}\label{subsec: calibration function}
After calculating the residual percentiles, the next step is to choose a specific quantile from the distribution to be saved in the matching cell of a node-prediction-window pair. To this end, we apply the trained quantile DNN to test samples from the second validation set $\chi_{2}$ to produce an initial PI for each test sample. Next, the residual quantiles, $Q^1_{i,j}, Q^2_{i,j}, \dots, Q^m_{i,j}$, are applied to the initial PI to obtain an adjusted PI one by one. By comparing the coverage and effective width of each resulting PI for each node-prediction-window combination against the true value of the validation set, we determine which residual provides the best performance. The optimal residual for the node-prediction-window combination is then recorded in the corresponding cell of the calibration table.

Specifically, our objective for selecting the residual to be recorded in the calibration table is to minimize: 
\begin{equation}
L = \mathop{\arg\min}\limits_{Q^m} \sum_{0}^{i} \sum_{0}^{j}  - \lambda \mathcal{C}{\textnormal{cov}} (Q^m_{i,j},\chi_{2}) + (1 - \lambda) \textnormal{PI} (Q^m_{i,j},\chi_{2}),
\label{eq:calibrations quantile loss}
\end{equation}
where $\lambda \in [0,1]$ is a weight to control the balance of $\textnormal{PI}$ and the coverage $\mathcal{C}{\textnormal{cov}}$ for a given node-prediction-window combination. Finally, the calibration table records the residual value where the index is the calibration quantile that gives the minimum $L$.

\subsection{Using the Trained DNN and Calibration Component}
Once we have trained the DNN model and the quantile function and constructed the calibration table as described above, we can then apply them to unseen test samples. This process is illustrated in Figure~\ref{fig:deployment}.

For a given test sample, the trained DNN generates a single-point prediction that is then passed to the quantile function to create the initial PI, $\{l, u\}$. Using the node id of the test sample and the prediction window, we search the calibration table for the corresponding residual value $\delta$. This value is used to amend the initial PI to $\{l-\delta, u+\delta\}$. A positive $\delta$ results in a wider PI, while a negative $\delta$ narrows it. It is possible that no residual value is available for a given combination of node id and prediction window, in which case the initial PI remains unchanged.

Residual errors of missing combinations of node ids and prediction windows can be later inserted into the calibration table. This can be done by using the true value measured after the prediction window to compute the residual errors of the node-prediction-window pair. Similarly, during deployment or every time the DNN model is re-trained, the calibration table can be also updated with the recorded predictions and true values. This way, the calibration table and quantile function can evolve over time to adapt to changes in the deployment environment.

\section{Evaluation Setup}
We evaluate \SystemName on real-world datasets. In particular, we apply \SystemName to representative DNN architectures for traffic forecasting and compare it with a wide range of baseline methods for uncertainty modeling. This section provides a detailed description of the datasets used in our evaluation, the baseline quantile methods that we compare \SystemName against, as well as the hardware and software platforms used for the experiments.

\subsection{Datasets}
\begin{table}[t!]
\centering
\scriptsize
\caption{Description of the Datasets Used.}
\begin{tabular}{ccccc}
\toprule
\textbf{Type}  & \textbf{Dataset}  & \textbf{No. sensors} & \textbf{Time period} &\textbf{Mean / Std} \\   
\midrule
\multirow{3}{*}{\rotatebox{90}{Speed}}
& METR-LA   & 207     & 03/01/2012 - 06/30/2012  & 53.71 / 20.26\\
& PeMS-BAY  & 325     & 01/01/2017 - 05/31/2017  & 62.61 / 9.59 \\
& PeMSD7(M) & 228     & 05/01/2012 - 06/30/2012  & 58.89 / 13.48\\

\cmidrule(lr){1-5}
\multirow{4}{*}{\rotatebox{90}{Flow}}
& PEMS03    & 358     & 09/01/2018 - 11/30/2018  & 179.3 / 143.7\\
& PEMS04    & 307     & 01/01/2018 - 02/28/2018  & 211.7 / 158.1\\
& PEMS07    & 883     & 05/01/2017 - 08/31/2017  & 308.5 / 188.2\\
& PEMS08    & 170     & 07/01/2016 - 08/31/2016  & 230.7 / 146.2\\

\bottomrule
\end{tabular}
\label{tab:dataset}
\end{table}

In our evaluation, we used seven public traffic datasets that are widely adopted in the literature. \cref{tab:dataset} provides the basics of each dataset, including the type of traffic data collected (e.g., speed or flow), the number of sensors used to collect the data, and the time period covered by each dataset. The raw data from these datasets are aggregated into 5-minute intervals, in align with previous literature, e.g., \cite{9346058}. 

\subsection{Base Traffic Forecasting Models}\label{subsec: base model}

To demonstrate the applicability of \SystemName over a variety of DNN-based traffic predictors, we adopt six representative models, namely, Spatio-Temporal Graph Convolutional Network (STGCN) \cite{Yu2018spatio}, Graph Wavenet (GWNet) \cite{ijcai_WuPLJZ19}, Graph Multi-Attention Network (GMAN) \cite{aaai_ZhengFW020}, Long-and Short-term Time-series Network (LSTNet) \cite{sigir_LaiCYL18}, and Multivariate Graph Neural Network (MTGNN) \cite{wu2020connect}. These models cover different variants of the graph neural network and  the long short-term memory (LSTM) architecture. 
They  have been widely used in the literature and represent 
different classes of deep learning-based methods for traffic forecasting. Note that our goal is to demonstrate the generalization ability of \SystemName to different models, rather than to compare the relative performance of the base DNN architectures. Therefor, our evaluation is designed to compare the performance of various quantile methods in uncertainty modeling, all within the same base DNN architecture.

\subsection{Competing Baselines}\label{subsec: competing baseline}
We compare \SystemName against five representative uncertainty quantification methods: 

\cparagraph{Historical Data-Based Methods} assume that traffic data, such as speeds or flow distribution, exhibit strong repeat patterns during the same period (e.g., the same day across weeks or the same hour across days). These methods predict current travel speed and flow distribution by utilizing the distribution of the same period from prior data, such as the distribution of the same day from the previous weeks. In our experiments, we consider two baseline methods that rely on historical data: \emph{Hist-D}, which uses data from the previous days in the training dataset for predictions, and \emph{Hist-W}, which uses the distribution of the same period from the previous weeks for prediction. For example, if we were to predict the PI for node $i$ at 8 a.m. on Monday, \emph{Hist-D} would calculate the mean $\mu_{i}^{t}$ and variance $\sigma_{i}^{t}$ across all previously seen samples of node $i$ at 8 a.m. of all Mondays. Once the mean and variance for each time slot and node of interest are calculated, these two values can be used to construct the prediction interval for that slot and node with $\{\mu_{i}^{t}$-$\sigma_{i}^{t}$, $\mu_{i}^{t}$+$\sigma_{i}^{t}\}$. Similarly, \emph{Hist-W} would compute the weekly mean and variances across four weeks of a month to compute the PI. 

\cparagraph{Bayesian uncertainty quantification} models the uncertainty in the model parameters using a likelihood function constructed by Bayesian modeling \cite{Blundell2015weight}. It also computes the data uncertainty by approximating the probability distribution over the model outputs through sampling and averaging over the resulting models. In our work, we use a Gaussian prior distribution with zero mean and unit variance \cite{Blundell2015weight}, and the MC sampling number of 50.

\cparagraph{Monte Carlo dropout} \cite{pmlr-v48-gal16} models predictive distributions by randomly switching off neurons in DNNs during testing. This generates different model outputs that can be interpreted as samples from a probabilistic distribution, allowing MC dropout to estimate the prediction probability. In our work, we added a dropout layer with a rate of 0.3 after the last hidden layer of the base traffic forecasting model and used a sample number of 50.

\cparagraph {Deep Quantile Regression (DQR)} can also estimate PIs \cite{Roger1978, nips_ChungNCS21}. Unlike conventional methods that minimize the averaged residual errors, DQR calculates the prediction errors at a specific quantile of the distribution. This method requires the use of differential models or tailored loss objectives such as NLL-based or pinball losses to generate the boundaries of the PI. In our quantile regression method, we use the 5th and 95th percentile estimates, following the configuration outlined in \cite{romano2019conformalized}.

\cparagraph{Conformal} prediction \cite{Mach2008Tutorial} is a post-processing method for quantifying prediction uncertainties. The main idea behind conformal prediction is to use a nonconformity measure to evaluate the similarity between new input data and the training data. Given a certain confidence level, conformal prediction constructs prediction regions that contain a certain fraction of predictions with the same nonconformity measure. In this study, we use inductive conformal prediction \cite{Papadopoulos08} due to its simplicity, which requires splitting the training data into two subsets. We use the same training dataset splitting ratio for conformal prediction and our approach.

\subsection{Experimental Setting}

\subsubsection{Hardware and software}
We implemented \SystemName as a Python package using PyTorch version 1.8.0. We conduct our model training and evaluation on a multi-core machine  equipped with a dual-socket 20-core, 2.50 GHz Xeon(R) Silver 4210 CPU, 256GB of DDR4 RAM, and 2x NVIDIA GeForce RTX 2080 Ti GPUs. Our system is running Ubuntu 18.04.5 with Linux kernel 4.15.0 and we execute the GPU code using CUDA version 11.1. 

\subsubsection{Training setup}\label{subsec: training}
For consistency with prior work \cite{DL2021Jiang}, we split the data into training, validation, and test sets at a ratio of 7:1:2. To model uncertainty, we further reserve 40\% of the training data for calibration (see \cref{subsec:Pre-processing}). We use the Kullback--Leibler divergence loss function to train Bayesian models \cite{Blundell2015weight}, and pinball loss for DQR, CQR, and the proposed \SystemName. For all other models, we use mean absolute error (MAE) as the loss function. We train all models using the Adam optimizer \cite{LeCun2015Adam} with a learning rate of 0.001, and a batch size of 64 (except for GMAN, where we use a batch size of 16 due to GPU memory constraints). The training process stops either after 200 training epochs or when the validation loss remaining unchanged for ten consistent epochs, following the standard practice in prior work \cite{DL2021Jiang}. 
 
To ensure a fair comparison, the hyperparameters of quantile-based methods are identical to those of CQR methods. For the additional hyperparameter in CQR, i.e., calibration miscoverage rate $\alpha_{\textnormal{cal}} $ controlling the movement and direction of the upper and lower marginal bounds, we use the calibration set to pick the best value.

\subsection{Evaluation Methodology}
We vary the observation step from 1 to 12, corresponding to historical data in the last 5 to 60 minutes, as each step contains sensor data over 5 minutes. We also set the prediction windows to $1, 2, \dots, 12$ steps by asking the based DNN model to predict traffic information in the next $5, 10, \dots, 60$ minutes. In other words, the base DNN model takes as input sensor data of the last $n$ minutes to predict information in the next $n$ minutes.

We evaluate the coverage and discrimination guarantees when applying an uncertainty method to each tested DNN model. As explained in Section \ref{sec:qm}, we use MPIW (where smaller values are better) to evaluate the discriminative performance and PICP (where larger values are better) to quantify the coverage of the generated PIs for each test sample. We compute PICP and MPIW by averaging the results across nodes the prediction window settings.

\section{Experimental Results}\label{sec:results}

We evaluate the effectiveness of \SystemName on seven datasets, comparing it to five state-of-the-art uncertainty modeling methods and two classical methods based on historical data methods.  We investigate the sensitivity of two hyperparameters: confidence level and the split between training and calibration data in \cref{subsec:Pre-processing}. Additionally, we conduct a detailed analysis to evaluate the impact of training data splitting ratio (\cref{sub:dsr}) and justify the calibration table construction method (\cref{sub:gvq}).

Highlights of our evaluation are: 
\begin{itemize}
\item \SystemName consistently delivers the best overall performance than the baseline uncertainty methods across base DNN architectures and datasets (\cref{sec:overall});
\item The adaptive calibration scheme introduced \SystemName gives a better trade-off between  the coverage and PI width for individual locations and sensor nodes over CQR (\cref{sec:api});
\item \SystemName gives more robust performance over CQR and DQR at different desired coverage rates (\cref{sub_different_confidence_level});
\item We showcase how \SystemName can be used to enhance the evaluation of traffic forecasting models (\cref{sec:petf}). 
\end{itemize}

\subsection{Overall Results\label{sec:overall}} 
\begin{table*}
\small
\centering
\caption {Performance Evaluation for Traffic Methods with Different Uncertainty Qualification Methods on Traffic Speed Datasets. \\($\mathrm{PICP}$ (\%) $\uparrow$ / $\mathrm{MPIW}$ $\downarrow$ )}
\begin{tabular}{lllllllccccccc}
\multicolumn{7}{l}{(a) METR-LA, Hist-D: $+$3.6\% / 34.2, Hist-W: $-$9.2\% / 28.2}\\ 
\toprule
Model Name& MC dropout & Bayesian & Conformal & DQR & CQR & \SystemName \\
\cmidrule(lr){1-7}
STGCN	&	$-$62.4\% /	3.0 	&	$-$44.9\% /	7.1 	&	$-$5.2\% /	13.7 	&	$+$0.5\% /	30.1 	&	$+$0.6\% /	20.6 	&	\textbf{$+$1.1\% /	20.5} 	\\
GWNet	&	$-$47.9\% /	3.5 	&	$-$38.9\% /	11.7 	&	$-$4.5\% /	11.9 	&	$-$0.3\% /	19.7 	&	$+$1.2\% /	15.6 	&	\textbf{$+$1.8\% /	15.1}	\\
MTGNN	&	$-$61.6\% /	2.4 	&	$-$63.0\% /	2.7 	&	$-$4.3\% /	11.8 	&	$-$0.1\% /	19.3 	&	$-$0.4\% /	14.7 	&	\textbf{$+$0.1\% /	14.6} 	\\
GMAN	&	$-$68.4\% /	1.7 	&	$-$11.9\% /	16.1 	&	$-$1.2\% /	18.1 	&	$+$0.9\% /	27.4 	&	$+$1.3\% /	20.4 	&	\textbf{$+$1.6\% /	20.1} 	\\
LSTNet	&	$-$54.7\% /	3.6 	&	$-$66.0\% /	3.0 	&	$-$2.7\% /	24.1 	&	$-$0.1\% /	29.0 	&	$+$0.3\% /	24.7 	&	\textbf{$+$0.4\% /	24.0} 	\\
\textbf{AVERAGE}	&	$-$59.00\% / 2.84 	&	$-$44.94\% /	8.12  	&	$-$3.58\% /	15.92  	&	$+$0.18	\% /	25.10 	&	$+$0.60\% /	19.20 	&	\textbf{$+$1.00\% /	18.86 } 	\\

\bottomrule

\multicolumn{7}{l}{} \\
\multicolumn{7}{l}{(b) PEMS-BAY, Hist-D: $-$13.8\% / 9.5, Hist-W: $-$32.7\% / 6.2}  \\ 
\toprule
  Model Name          & MC dropout & Bayesian & Conformal & DQR & CQR & \SystemName \\
\cmidrule(lr){1-7}
STGCN	&	$-$58.8\% /	1.6 	&	$-$38.1\% /	3.7 	&	$-$3.3\% /	6.8 	&	$-$2.4\% /	9.8 	&	$-$0.1\% /	8.2 	&	\textbf{$+$0.0\% /	8.1} 	\\
GWNet	&	$-$47.3\% /	2.9 	&	$-$32.9\% /	3.9 	&	$-$1.8\% /	5.9 	&	$-$0.9\% /	7.8 	&	$+$0.1\% /	6.3 	&	\textbf{$+$0.6\% /	6.2} 	\\
MTGNN	&	$-$66.4\% /	1.2 	&	$-$67.5\% /	1.2 	&	$-$1.6\% /	5.8 	&	$-$0.6\% /	7.8 	&	$-$0.2\% /	6.2 	&	\textbf{$+$0.2\% /	6.1} 	\\
GMAN	&	$-$64.8\% /	1.4 	&	$-$3.3\% /	7.3 	&	$-$5.5\% /	5.7 	&	$+$0.5\% /	9.7 	&	$-$0.2\% /	7.6 	&	\textbf{$+$0.5\% /	7.6} 	\\
LSTNet	&	$-$31.9\% /	29.2 	&	$-$84.0\% /	0.3 	&	$-$0.6\% /	12.3 	&	$-$5.1\% /	10.5 	&	$-$1.5\% /	10.3 	&	\textbf{$-$0.9\% /	9.9} 	\\
\textbf{AVERAGE}	&	$-$53.84\% / 7.26  	&	$-$45.16\% /	3.28 	&	$-$2.56\% /	7.30	&	$-$1.70\% /	9.12 	&	$-$0.38\% /	7.72	&	\textbf{$+$0.08\% /	7.58} 	\\
\bottomrule

\multicolumn{7}{l}{} \\
\multicolumn{7}{l}{(c) PEMSD7M, Hist-D: $-$15.3\% / 10.7, Hist-W: $-$35.0\% / 7.9}  \\ 

\toprule
  Model Name          & MC dropout & Bayesian & Conformal & DQR & CQR & \SystemName  \\
\cmidrule(lr){1-7}
STGCN	&	$-$56.0\% /	2.8 	&	$-$36.7\% /	5.8 	&	$-$7.1\% /	12.2 	&	$-$2.4\% /	17.0 	&	+0.4\% /	17.0 	&	\textbf{$+$1.1\% /	16.9} 	\\
GWNet	&	$-$40.1\% /	3.6 	&	$-$37.2\% /	6.6 	&	$-$4.6\% /	9.6 	&	$-$0.5\% /	14.4 	&	$-$0.2\% /	13.0 	&	\textbf{$+$0.1\% /	12.9} 	\\
MTGNN	&	$-$37.8\% /	4.6 	&	$-$53.2\% /	2.8 	&	$-$2.2\% /	11.6 	&	$-$1.0\% /	13.0 	&	$-$0.2\% /	13.0 	&	\textbf{$+$0.2\% /	12.9} 	\\
GMAN	&	$-$51.8\% /	2.7 	&	$-$2.1\% /	14.4 	&	$-$3.7\% /	12.2 	&	$-$0.7\% /	17.0 	&	$+$0.7\% /	15.4 	&	\textbf{$+$0.8\% /	15.1} 	\\
LSTNet	&	$-$33.0\% /	5.9 	&	$-$16.2\% /	9.6 	&	$-$3.1\% /	17.7 	&	$-$4.8\% /	16.7 	&	$-$0.4\% /	17.6 	& \textbf{$+$0.0\% /	17.6} 	\\
\textbf{AVERAGE}	&	$-$43.74\% / 3.92  	&	$-$29.08\% /	7.84  	&	$-$4.14\% /	12.66  	&	$+$1.88\% /	15.62 	&	$+$0.06\% /	15.20 	&	\textbf{$+$0.44\% /	15.08 } 	\\

\bottomrule
\label{table_res_speed}
\end{tabular}
\end{table*}

\begin{table*}
\small
\centering
\caption{Performance Evaluation for Traffic Methods with Different Uncertainty Qualification Methods on Traffic Flow Datasets\\ ($\mathrm{PICP}$ (\%) $\uparrow$ / $\mathrm{MPIW}$ $\downarrow$)}
\begin{tabular}{lllllllccccccc}

\multicolumn{7}{l}{(a) PEMS03, Hist-D: $-$27.4\% / 67.6, Hist-W: $-$56.9\% / 37.2 }  \\ 
\toprule
  Model Name          & MC dropout & Bayesian & Conformal & DQR & CQR & \SystemName  \\
\cmidrule(lr){1-7}
STGCN	&	$-$54.9\% /	18.0 	&	$-$18.1\% /	57.9 	&	$-$3.8\% /	75.8 	&	$-$10.8\% /	61.8 	&	$-$2.1\% /	81.6 	&	\textbf{$-$0.5\% /	78.7} 	\\
GWNet	&	$-$36.0\% /	30.4 	&	$-$8.2\% /	79.6 	&	$-$1.5\% /	64.3 	&	$-$4.5\% /	55.3 	&	$+$1.5\% /	66.2 	&	\textbf{$+$1.2\% /	66.1} 	\\
MTGNN	&	$-$27.3\% /	40.9 	&	$-$49.9\% /	18.8 	&	$-$0.8\% /	65.7 	&	$-$2.0\% /	58.6 	&	$-$0.5\% /	60.6 	&	\textbf{$-$0.4\% /	61.6} 	\\
GMAN	&	$-$41.8\% /	29.8 	&	$+$5.7\% /	127.1 	&	$-$1.5\% /	64.3 	&	$-$8.5\% /	67.8 	&	$-$1.2\% /	85.5 	&	\textbf{$-$1.3\% /	85.9} 	\\
LSTNet	&	$-$31.9\% /	29.2 	&	$-$57.8\% /	40.4 	&	$-$2.3\% /	73.2 	&	$-$4.0\% /	63.3 	&	$+$0.4\% /	70.8 	&	\textbf{$+$0.4\% /	69.1} 	\\
\textbf{AVERAGE}	&	$-$38.38\% /	29.66  	&	$-$25.66\% /	64.76  	&	$-$1.98\% /	68.66 	&	$-$5.96\% /	61.36  	&	$-$0.38\% /	72.94  	&	\textbf{$-$0.12\% /	72.28 } 	\\

\bottomrule

\multicolumn{7}{l}{} \\
\multicolumn{7}{l}{(b) PEMS04, Hist-D: $-$21.8\% / 92.2, Hist-W: $-$58.2\% / 52.1}  \\ 
\toprule
  Model Name          & MC dropout & Bayesian & Conformal & DQR & CQR & \SystemName  \\
\cmidrule(lr){1-7}
STGCN	&	$-$51.5\% /	22.6 	&	$-$19.4\% /	64.1 	&	$+$0.9\% /	108.7 	&	$-$4.1\% /	86.7 	&	$-$0.3\% /	93.5 	&	\textbf{$+$0.2\% /	94.7} 	\\
GWNet	&	$-$39.0\% /	37.5 	&	$-$3.1\% /	115.6 	&	$+$0.5\% /	92.5 	&	$-$1.3\% /	84.4 	&	$+$0.9\% /	85.5 	&	\textbf{$+$1.1\% /	85.3} 	\\
MTGNN	&	$-$28.1\% /	55.0 	&	$-$46.8\% /	25.6 	&	$+$0.5\% /	91.3 	&	$-$1.4\% /	81.2 	&	$+$0.5\% /	81.2 	&	\textbf{$+$0.8\% /	82.4} 	\\
GMAN	&	$-$48.6\% /	32.0 	&	$-$0.2\% /	116.4 	&	$+$2.0\% /	117.3 	&	$+$5.6\% /	132.7 	&	$+$0.2\% /	97.9 	&	\textbf{$+$0.9\% /	100.9} 	\\
LSTNet	&	$-$29.9\% /	39.4 	&	$-$52.6\% /	25.4 	&	$+$0.8\% /	102.8 	&	$-$0.3\% /	92.8 	&	$+$1.3\% /	93.1 	&	\textbf{$+$1.4\% /	92.3} 	\\
\textbf{AVERAGE}	&	$-$39.42\% /	37.30  	&	$-$24.42\% /	69.42  	&	$+$0.94\% /	102.52  	&	$-$0.30\% /	95.56  	&	$+$0.52\% /	90.24  	&	\textbf{$+$0.88\% /	91.12 } 	\\
\bottomrule

\multicolumn{7}{l}{} \\
\multicolumn{7}{l}{(c) PEMS07, Hist-D: $-$20.7\% / 110.2, Hist-W: $-$21.2\% / 59.6}  \\ 
\toprule
  Model Name          & MC dropout & Bayesian & Conformal & DQR & CQR & \SystemName  \\
\cmidrule(lr){1-7}
STGCN	&	$-$51.9\% /	23.9 	&	$-$18.4\% /	88.4 	&	$-$1.3\% /	116.3 	&	$-$8.4\% /	94.9 	&	$-$0.5\% /	112.2 	&	\textbf{$+$0.0\% /	108.7} 	\\
GWNet	&	$-$80.6\% /	5.9 	&	$-$6.8\% /	117.9 	&	$-$1.3\% /	93.5 	&	$-$1.1\% /	92.9 	&	$+$0.4\% /	94.9 	&	\textbf{$+$0.4\% /	92.5} 	\\
MTGNN	&	$-$52.6\% /	24.8 	&	$-$51.8\% /	23.2 	&	$-$2.6\% /	88.3 	&	$-$1.7\% /	88.1 	&	$-$0.3\% /	90.7 	&	\textbf{$-$0.1\% /	89.1} 	\\
GMAN	&	$-$45.2\% /	35.9 	&	$+$0.7\% /	145.6 	&	$-$6.5\% /	137.1 	&	$+$4.5\% /	154.7 	&	$-$1.8\% /	117.4 	&	\textbf{$-$1.8\% /	116.8}	\\
LSTNet	&	$-$52.6\% /	24.8 	&	$-$53.1\% /	30.3 	&	$-$2.2\% /	107.8 	&	$-$1.4\% /	107.9 	&	$-$0.3\% /	109.2 	&	\textbf{$+$0.0\% /	107.1} 	\\
\textbf{AVERAGE}	&	$-$56.58\% / 23.06  	&	$-$25.88\% /	81.08  	&	$-$2.78\% /	108.60 	&	$-$1.62\% /	107.70 	&	$-$0.50\% /	104.88  	&	\textbf{$-$0.30\% /	102.84 } 	\\

\bottomrule

\multicolumn{7}{l}{} \\
\multicolumn{7}{l}{(d) PEMS08, Hist-D: $-$36.4\% / 80.2, Hist-W: $-$29.8\% / 41.9}  \\ 
\toprule
  Model Name          & MC dropout & Bayesian & Conformal & DQR & CQR & \SystemName \\
\cmidrule(lr){1-7}
STGCN	&	$-$50.8\% /	18.8 	&	$-$15.8\% /	65.4 	&	$-$0.4\% /	90.6 	&	$-$5.9\% /	74.4 	&	$-$0.2\% /	89.0 	&	\textbf{$+$0.1\% /	83.4} 	\\
GWNet	&	$-$37.1\% /	36.3 	&	$-$1.3\% /	99.8 	&	$-$0.7\% /	70.6 	&	$+$0.6\% /	74.8 	&	$+$0.1\% /	75.1 	&	\textbf{$+$0.1\% /	72.0} 	\\
MTGNN	&	$-$26.9	\% /	47.4 	&	$-$35.3\% /	27.8 	&	$-$1.3\% /	69.1 	&	$-$1.2\% /	68.5 	&	$+$1.4\% /	73.1 	&	\textbf{$+$1.2\% /	71.1} 	\\
GMAN	&	$-$39.3	\% /	31.0 	&	$+$0.9\% /	104.7 	&	$-$2.1\% /	118.7 	&	$+$1.2\% /	92.6 	&	$+$3.0\% /	93.0 	&	\textbf{$+$3.4\% /	92.6} 	\\
LSTNet	&	$-$36.1	\% /	28.2 	&	$+$1.0\% /	145.0 	&	$+$0.5\% /	85.1 	&	$-$2.0\% /	77.3 	&	$+$0.7\% /	80.3 	&	\textbf{$+$0.8\% /	78.3} 	\\
\textbf{AVERAGE}	&	$-$38.04\% /	32.34  	&	$-$10.10\% /	88.54 	&	$-$0.80\% /	86.82  	&	$-$1.46\% /	77.52  	&	$+$1.00\% /	82.10  	&	\textbf{$+$1.12\% /	79.48 } 	\\

\bottomrule
\label{table_res_flow}
\end{tabular}
\end{table*}

In this experiment, we set an expectation of a 90\% coverage rate, and we evaluate if the quantify method can reach or even exceed the 90\% coverage rate across test samples using PICP.  Later in \cref{sub_different_confidence_level}, we compare the performance under different coverage rate expectations. 

Tables \ref{table_res_speed} and \ref{table_res_flow} present the results obtained when applying each evaluated uncertainty method to different DNN models and datasets. The best-performing results are highlighted in bold for each base DNN model. In the corresponding column of each uncertainty method, we first show the improvement of PICP (as a percentage with respect to a 90\% coverage rate), followed by the MPIW that measures the width of the PI (\emph{lower-is-better}). In this context, a positive PICP improvement means that the PI's coverage exceeds the expectation of a 90\% coverage rate, while a negative PCIP improvement ($-$) means that the PI's coverage falls below the expectation.  For example, in Table \ref{table_res_speed}(a), applying \SystemName to STGCN yields $+$1.1\% / 20.5. This should read as that \SystemName achieves a 91.1\% coverage rate (above the coverage expectation) with 20.5 in MPIW (i.e., the difference between the upper and the lower bound of the PI is 20.5). For each dataset, we report the average PCIP and MPIW across test samples. 

As expected, the MPIW given by all methods depends on the base DNN model's capability, because a larger MPIW (or a wider PI) is needed to ensure the coverage for a less accurate DNN model. However, we observe that \SystemName achieves or exceeds the desired 90\% coverage rate for most test cases. For a handful of cases where the \SystemName PICP falls below 90\%, the resulting coverage rate remains close to the target of 90\%, with a coverage rate of at least 88\% (i.e., $\geqslant -1.8\%$ in the tables). In addition to achieving a good coverage rate, \SystemName produces small MPIW, thus providing a meaningful uncertainty measure for decision-making.

Compared to \SystemName, non-frequentist baselines such as MC dropout, Bayesian, and Conformal methods provide a narrow PI, resulting in a small MPIW. However, they struggle to meet the coverage expectation, where the true value of the test sample often falls outside the PI. This suggests that the PI given by non-frequentist baselines can mislead decision-making by being too optimistic or over-conservative for traffic forecasting. The performance of non-frequentist baselines is also not consistent, showing significant variance depending on the test datasets and DNN models. For instance, while Bayesian gives a coverage rate of 86.89\% on PEMS-BAY when using GMAN as the base DNN, its coverage rate is only 22.52\% on the same dataset when using MTGNN as the base model. 

By utilizing previously collected data, Hist-D gives good coverage but with a large MPIW, leading to over-conservative forecasts. In contrast, quantile-based methods such as DQR and CQR significantly outperform others by providing a better coverage guarantee. By incorporating an adaptive scheme to adjust the initial PI based on the test node and prediction window, \SystemName improves upon CQR and DQR with higher PCIP and smaller MPIW for all test cases. \SystemName also delivers consistent good performance across datasets and base DNN models, demonstrating the robustness of our approach. This further highlights the advantage of quantile-based methods over non-frequentist baselines, which exhibit inconsistent and varying performance depending on the test datasets and base DNN models. 

\subsection{Adaptive PI Adjustments\label{sec:api}}
\begin{figure}
  \centering
  \includegraphics[width=0.95\linewidth]{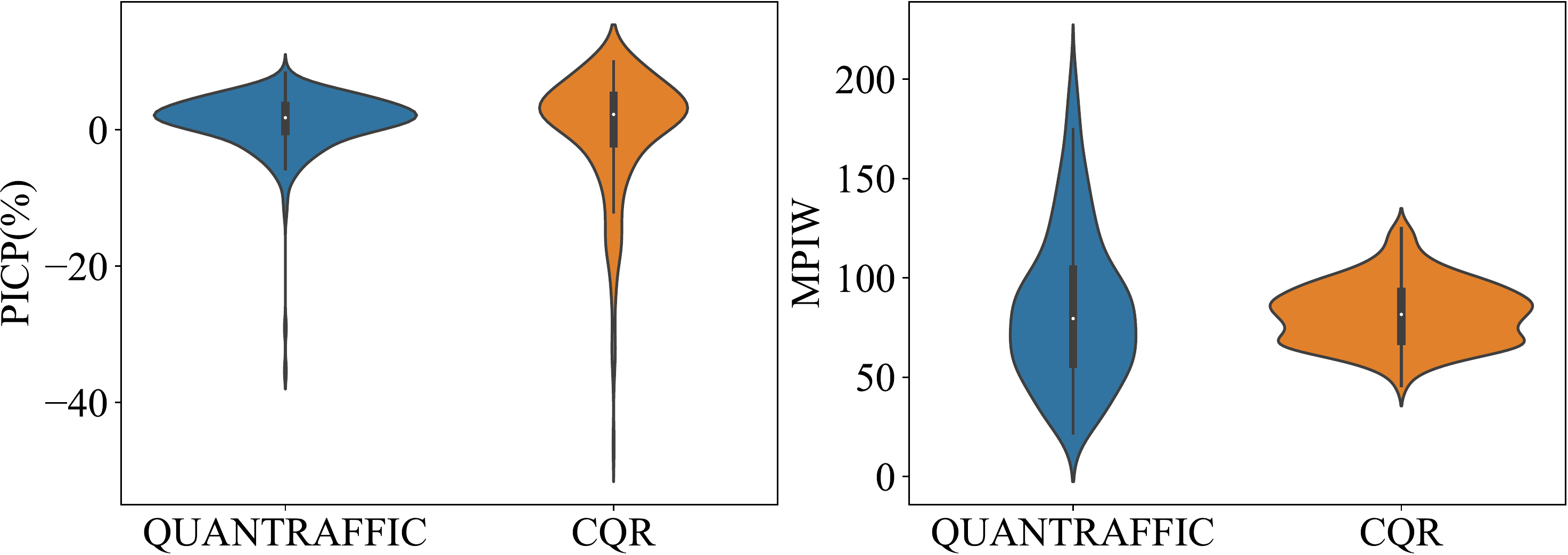}
  \caption{Violin diagrams show PICP and MPIW by applying CQR and \SystemName to GWNet model tested on the PEMS04.  The thick black line shows where 50\% of the data locates.}
  \label{fig:nodeqca}
\end{figure}
\SystemName advances CQR by adaptively adjusting the initial PI through a dedicated calibration table that uses different adjustments rather than a constant global value adopted by standard CQR. To demonstrate the benefit of \SystemName over CQR, we closely examine the test samples of the PEMS04 dataset when using GWNet  as the base DNN for traffic forecasting. The results are given in the violin diagram of \cref{fig:nodeqca}. In the diagram, the width of the violin indicates the density of test data points with a given PICP, and the center of the plot is a box plot with the median and quartiles the test samples fall into a specific PICP (or coverage rate). 

For this test scenario, the averaged PICP given by CQR and \SystemName is comparable at 90.9\% and 91.1\%, respectively. However, the effectiveness of the PI can vary significantly for individual nodes, highlighting the importance of having an adaptive scheme. Upon closer inspection of \cref{fig:nodeqca}, we see that the coverage rates provided by \SystemName are more uniformly located around the desired coverage rate of 90\% across nodes. In contrast, the coverage rate provided by CQR is highly diverse, ranging from 45\% to 100\%. In other words, CQR's performance is less consistent and less robust than \SystemName, as CQR can lead to a poor coverage guarantee for individual nodes under a given prediction window.

\subsection{Case Study of Selected Test Samples}
\begin{figure*}[t!]
    \centering
    \includegraphics[width=0.9\linewidth]{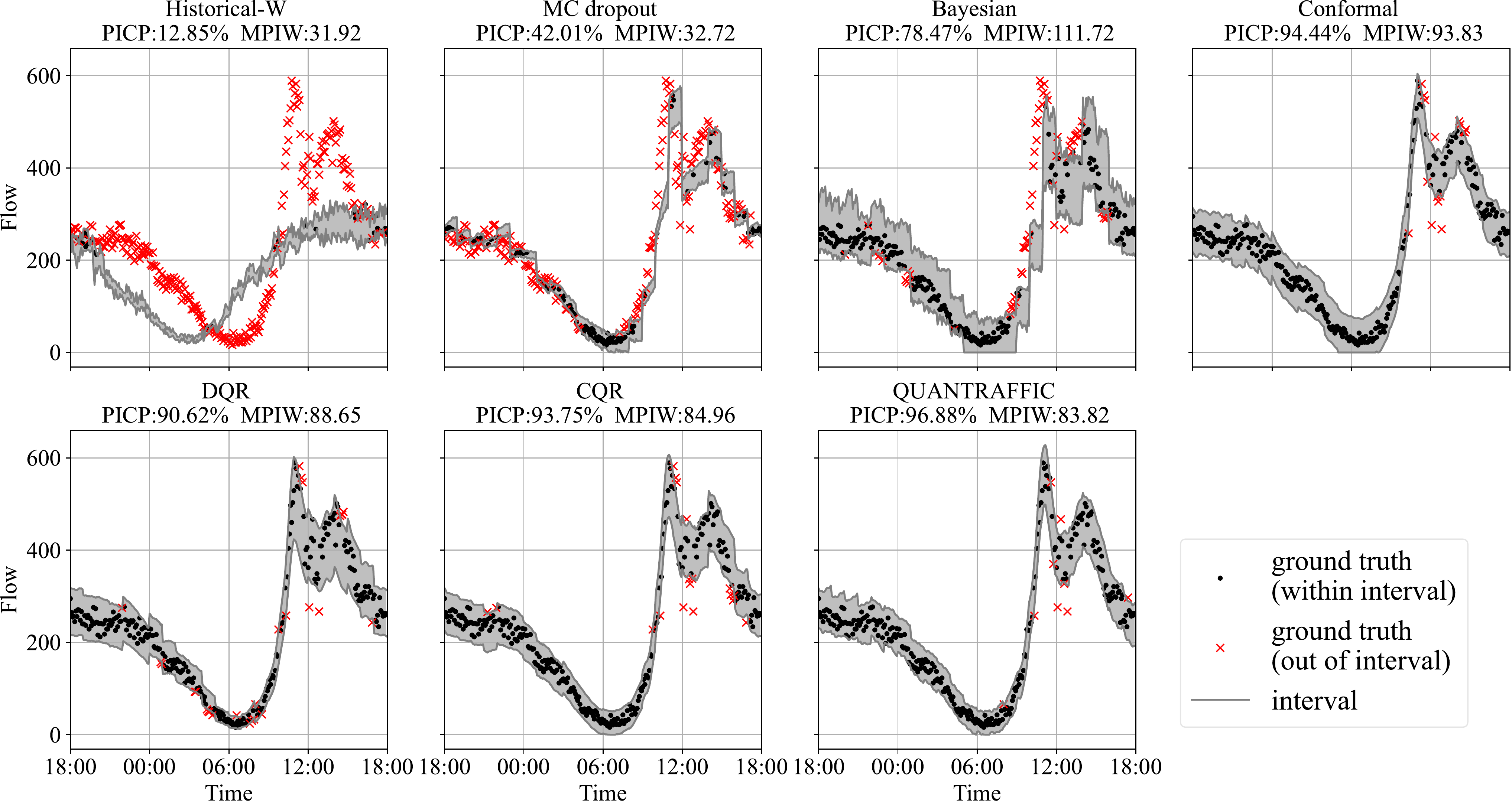}
    \caption{
      Coverage and prediction intervals for different uncertainty quanlification models of PEMS04 from 2012-02-18 18:00 to 2012-02-19 18:00. }
      \label{fig_interval}
\end{figure*}
In \cref{fig_interval}, we closely examine the PI generated by different uncertainty methods for traffic flow prediction performed on PEMS04 on the date from 2012-02-18 18:00 to 2012-02-19 18:00. In the diagram, the PI is represented as a grey area, while the ground truth of a test sample is represented by a point. If a point falls within the grey belt, the generated PI covers the true value, otherwise, it fails to cover the true value. The grey area of a good strategy should cover as many points as possible while being as narrow as possible.

Methods like MC drop and Hist-W result in a small grey area with a small MPIW, but their PIs fail to cover the ground truth for a large number of test samples, leading to a poor PICP and insufficient coverage. Quantile methods like DQR and CQR perform significantly better than MC dropout and Bayesian. Compared to DQR and CQR, \SystemName produces PIs that cover the ground truth of more test samples (i.e., a higher PICP) with a smaller MPIW (i.e., a narrower grey area in the diagram). This example demonstrates the effectiveness of our adaptive scheme.

\subsection{Impact of Desired Coverage Rates}\label{sub_different_confidence_level}

\begin{table}[t]
  \centering
  \caption{Performance Comparison of Quantile Methods under Different Coverage Levels for $\mathrm{PICP}$ (\%) $\uparrow$ / $\mathrm{MPIW}$ $\downarrow$.}
  \begin{tabular}{rllll}
    \toprule
    \begin{tabular}[x]{@{}c@{}} Expected \\ Coverage \end{tabular} &  DQR  & CQR & \SystemName\\
    \cmidrule(lr){1-4}
      0.6	& $-$41.1\% / 1.0 & $-$0.1\% / 4.1 &	\textbf{$+$0.1\%	/ 4.1} \\
      0.7	& $-$33.0\% / 2.0 & $-$1.5\% / 5.3 &	\textbf{$-$1.1\%	/ 5.2} \\
      0.8	& $-$21.2\% / 4.3 & $-$1.7\% / 7.8 &	\textbf{$-$1.4\%	/ 7.6} \\
      0.95	& $-$5.2\%	/ 21.5 &	$-$8.5\% / 20.9 &	\textbf{$-$2.5\% / 20.2} \\
      \textbf{Avg.}	& $-$30.09\%	/ 5.76	& $-$2.50\%	/ 8.15 &	\textbf{$-$1.11\%	/ 7.97}\\
    \bottomrule
    \label{tab: different quantile}
  \end{tabular}
\end{table}
So far, our evaluation set 90\% as the targeting coverage rate. In this section, we investigate the impact of the desired coverage rate on the performance of uncertainty methods. Previous evaluations have shown that DQR, CQR and \SystemName are the best-performing methods. Therefore, we focus on quantile-based methods in this experiment. We apply the quantile methods to the METR-LA dataset using GWNet as the base DNN model.  We vary the expected coverage level, $1-\alpha$, from 0.6 to 0.95, corresponding to a target coverage rate of 60\% to 95\%, respectively, to evaluate the usefulness of traffic uncertainty models in different practical scenarios.  

\cref{tab: different quantile} compares the PCIP/MPIW given by each quantile method averaged across the test samples for a target coverage level. As expected, as we increase the coverage level to ensure a coverage rate, a larger MPIW (and a wider PI) is required. Once again, \SystemName outperforms the other methods, providing the best coverage rate across settings. While DQR has the smallest MPIW in most cases, it gives a poor coverage rate, which can be up to 41\% below the expected value. This suggests that the PI provided by DQR is too narrow (and hence has a small MPIW) to cover the true value of the prediction. CQR addresses the issues of DQR by using a constant adjustment value. However, it can still provide poor coverage or a PI that is too wide for some individual nodes. By dynamically adjusting the PI based on individual nodes, \SystemName provides the best overall performance. This further demonstrates the effectiveness and flexibility of our approach in adapting to different sensor nodes or locations. 

\subsection{Impact of Training-Calibration Split Ratio}\label{sub:dsr}
\begin{table}
  \centering
  \caption{Performance Comparison of Different Training-calibration Ratio $\beta$ on Traffic Flow Dataset PEMS08 ($\mathrm{PICP}$ (\%) $\uparrow$ / $\mathrm{MPIW}$ $\downarrow$)} 
  \begin{tabular}{ccc}
    \toprule
    $\beta$ & CQR & \SystemName \\
    \cmidrule(lr){1-3}
    1:6 & $+$2.4\% / \textbf{88.7} & \textbf{$+$3.0\%} / 88.9 \\
    2:5 & $+$0.5\% / 76.2 & \textbf{$+$0.8\% / 76.1} \\
    3:4 & $+$0.1\% / 72.5 & \textbf{$+$0.1\% / 72.3} \\
    1:1 & $+$0.3\% / 74.6 & \textbf{$+$0.5\% / 73.6} \\
    4:3 & $+$2.6\% / 73.9 & \textbf{$+$2.6\% / 71.7} \\
    5:2 & $+$1.4\% / 73.6 & \textbf{$+$1.6\% / 73.4} \\
    6:1 & $+$2.1\% / \textbf{74.3} & \textbf{$+$2.4\%} / 75.1 \\
    \textbf{Avg.} & $+$1.35\% / 76.25 & \textbf{$+$1.56\% / 75.88}\\ 
    \bottomrule
  \end{tabular}
  \label{sub_different_splitting_rate}
\end{table}
To build the calibration component, \SystemName and CQR require setting aside some data from the training dataset as the calibration data. Our experiments described so far leave 40\% of the original training data as the calibration dataset. In this experiment, we evaluate how the training-calibration dataset ratio affects the performance on \SystemName and CQR. Specifically, We vary the ratio ($\beta$) between the DNN model training data and the calibration data to examine the impact on performance. Specifically, we evaluate the ratios of 6:1, 5:2, 4:3, 1:1, 3:4, 2:5, and 1:6. We then apply \SystemName and CQR to the PEMS08 dataset, using GWNet as the base DNN model, with a target coverage rate of 90\%.

The result is given in \cref{sub_different_splitting_rate}. Leaving too few samples to train the base DNN model may lead to decreased model accuracy, resulting in a wider PI (and larger MPIW) required to ensure sufficient coverage. For instance, when the training-calibration data ratio $\beta$ is set to 1:6 or 2:5, both CQR and \SystemName produce a wide PI. However, insufficient calibration data can also impact the accuracy of the uncertainty model, given that \SystemName relies on partitioning the calibration dataset to construct the calibration table. As a result, a smaller calibration dataset (e.g., when $\beta$ is 6:1) can affect the performance of the model. Nevertheless, for most experimental settings, \SystemName outperforms CQR in terms of both the PICP and MPIW metrics.

\subsection{Quantile Search for Calibration Table Construction} \label{sub:gvq}
\begin{table}[t!]
  \centering
  \caption{Performance Comparison of Grid Search and \SystemName \\ ($\mathrm{PICP}$ (\%) $\uparrow$ / $\mathrm{MPIW}$ $\downarrow$)}
  \begin{tabular}{rrr}
    \toprule
    Frequency & Grid search	&	\SystemName \\
    \midrule
    10 &	$-$49.1\%	/ 31.5 &	\textbf{$+$0.9\%	/	73.1 }\\
    20 & $-$23.8\%	/	44.7 & \textbf{$+$0.7\%	/	72.0} \\
    40 & $-$9.9\%	/	56.3 & \textbf{$+$0.5\%	/	71.4} \\
    60 & $-$5.6\%	/	61.9 & \textbf{$+$0.4\%	/	71.1} \\
    80 & $-$3.7\%	/	64.5 & \textbf{$+$0.4\%	/	70.9} \\
    100 &	$-$2.7\%	/	66.3 & \textbf{$+$0.4\%	/	70.9} \\
    \bottomrule
  \end{tabular}
  \label{sub_grid_vs_quantile}
\end{table}

As described in Section \ref{subsec: component}, to construct the calibration table, we first apply the trained DNN and quantile function to the calibration dataset to obtain the prediction residuals. We then partition the residuals into continuous intervals with equal percentiles rather than predefined equal-sized intervals. In this experiment, we compare our quantile-based approach against a grid method for projecting the residuals onto intervals. The results were obtained on the PEMS08 dataset using GWNet as the base DNN model and a target coverage rate of 90\%. The results are given in \cref{sub_grid_vs_quantile}, which demonstrates the effectiveness of quantile search over the grid search method.

Using quantiles over a specific range can be advantageous in situations where traffic data is partially missing, and the range of possible values for residuals is not well-defined. Quantile search reduces the need for dense interval steps by focusing the search on the most promising regions in the parameter space, thereby increasing the efficiency of the search process. It can also help avoid overfitting and increase robustness to outliers. On the other hand, in grid search, missing traffic data can cause a narrow range of possible values, requiring more interval steps to capture better adjustments.

\subsection{Performance Evaluation of Traffic Forecasting Models\label{sec:petf}}
\begin{table}[t!]
  \centering
  \caption{Performance Evaluation Based on Point and Uncertainty Metrics}
  \label{tab: point vs uncer}
  \scriptsize
  \begin{tabular}{clrrrrr}
  \toprule
  \multirow{2}*[\multirowoffset]{Data} & \multirow{2}*[\multirowoffset]{Method} & \multicolumn{3}{c}{Point estimation} & \multicolumn{2}{c}{Uncertainty estimation} \\
  \cmidrule(lr){3-5}\cmidrule(lr){6-7}
  & & MAPE $\downarrow$ & RMSE $\downarrow$ & MAE $\downarrow$ & PICP $\uparrow$ & MPIW $\downarrow$ \\
  \midrule
  \multirow{5}*{\rotatebox{90}{METR-LA}}
  & STGCN  & 13.09\% & 11.81 & 5.08 &	90.13\% & 19.52 \\
  & GWNet  & \textbf{11.93\%} & \textbf{11.49} & \textbf{4.85} & 90.14\% & 15.22 \\
  & MTGNN  & 12.17\% & 11.62 & 4.87 &	\textbf{90.24\%} & \textbf{14.79} \\
  & GMAN   & 14.10\% & 11.91 & 5.48 &	89.36\%	& 17.27 \\
  & LSTNet & 15.38\% & 11.89 & 6.34 &	89.83\%	& 23.28 \\
  \cmidrule(lr){1-7}
  \multirow{5}*{\rotatebox{90}{PEMS08}}
  & STGCN  & 11.15\% & 18.01 & 28.11 & 90.06\% & 83.44 \\
  & GWNet  & \textbf{9.59\%} & \textbf{15.05} & \textbf{23.81} & 90.05\% & 72.01  \\
  & MTGNN  & 9.69\% & 15.15 & 23.98 & \textbf{91.16\%} & \textbf{71.13} \\
  & GMAN   & 12.99\% & 18.29 & 28.06 & 90.20\% & 80.77 \\
  & LSTNet & 10.86\% & 17.07 & 26.83 & 90.78\% &  78.31 \\
  \bottomrule
  \end{tabular}
\end{table}

The ability to model prediction uncertainty can also be useful in evaluating the creditability of a traffic forecasting mode. In this experiment, we extend our evaluation to explore the trade-off between accuracy and confidence in traffic forecasting tasks. There is increasing research effort in discovering the best-scoring traffic predictor in specific or ideal scenarios \cite{Vlahogianni2021}. Meanwhile, the reliability of the experimental evaluation is often neglected. For example, a traffic forecasting model could perform well on average in point estimation matrices but be less accurate during peak hours than at night. In this evaluation, we apply \SystemName to METR-LA and PEMS08 and compare the prediction accuracy measured by commonly used loss functions, Mean Absolute Error (MAE), Root Mean Square Error (RMSE), Mean Absolute Percentage Error (MAPE) and the corresponding coverage and discrimination on test data measured in PICP and MPIW. 

As can be seen from the \cref{tab: point vs uncer}, GWNet would be regarded as the most accurate model using traditional point-based evaluation metrics, but it is less reliable in coverage and discrimination metrics. While evaluating model credibility is not the focus of this work, our approach can provide a new measure for the performance evaluation of traffic forecasting models. 

\section{Discussion}

Naturally, there is room for future work and improvement. We discuss a few points here.

\cparagraph{Alternative quantile functions.} In the quantile DNN model training (see \cref{subsec:Quantile}), we attach a linear layer  to the last layer of the DNN and use the linear layer as the quantile function. An alternative could be using a dedicated network to generate two separate predictions. In this way, for each input, the DNN model would produce two separate outputs from the modified last layer, one for the lower bound and the other for the upper bound. The pinball loss assigns orthogonal weights to both predictions and uses them to calculate the loss. We leave this as our future work.

\cparagraph{Calibration component.} In this paper, we use a dedicated calibration table to provide customized PI adjustments for individual nodes. Another way of doing this is to train a calibration function using, e.g., linear regression or a neural network. Doing so would require having sufficient training data to learn an accurate calibration function. 

\cparagraph{Coverage goal.} \SystemName aims to produce a PI to meet the user-defined coverage level. In general, a higher coverage level guarantees stronger prediction accuracy  In practise usage, a high coverage level can be employed to ensure, for instance, the worst-case arrival time. In contrast, a lower coverage level may suffice if the user is willing to tolerate a certain level of prediction error in traffic information, for example, if the consequence of missing an event is insignificant. As such, techniques for learning and modeling user needs \cite{Webb2001} are complementary to \SystemName. 

\cparagraph{Other calibration techniques.}
\SystemName employs a straightforward yet efficient method of utilizing a calibration table to adjust the PI. Additionally, other post-processing techniques can be used to enhance initial predictions. For instance, ensemble methods \cite{nips_Lakshminarayanan17} can leverage multiple prediction models, and data augmentation techniques \cite{thulasidasan2019mixup} involve applying various data augmentation methods to the test input to obtain multiple predictions to form the PI. Our future work will investigate the use of these techniques. 

\cparagraph{Data distribution drifts.} CQR leverages conformal prediction for uncertainty measurement, which is based on the fundamental assumption of data exchangeability \cite{Jing2018}. Exchangeability refers to the statistical properties of the data remaining unchanged regardless of the order in which it is presented. However, in many real-world applications, particularly in traffic forecasting, the data distribution may be shifted due to various factors such as changes in road infrastructure (e.g., temporary road closures), weather patterns, and traffic patterns (e.g., traffic accidents), leading to a shift in the test data distribution with respect to the training data. Our future work will investigate how online  adaptation techniques can be employed to detect data drift and mitigate the issue of data drift through (e.g., continuous learning \cite{9349197}).

\cparagraph{Calibration data size.}
\SystemName and CQR require setting aside some data from the model training dataset as the calibration data.  If the calibration set is not representative of the test set, the performance of CQR and \SystemName may suffer, as shown in \cref{sub_different_confidence_level}. As such, techniques for data augmentation \cite{ijcai2021p631} like  basic data augmentation methods (e.g., window cropping \cite{le2016data}), Deep Generative Models \cite{8903012} and data selection \cite{eetemadi2015survey} like active learning  and scoring functions \cite{eck2005low} are orthogonal to our approach.

\section{Conclusions}\label{sec:conclu}
We have presented \SystemName, a framework that enhances the capability of DNN-based traffic forecasting models to quantify the uncertainty of their predictions. Specifically, \SystemName generates upper and lower bounds of the prediction, which is useful in tasks like emergency route planning to ensure the worst-case arrival time. Our framework is generic, applicable to any DNN model, and does not alter the DNN model's underlying architecture during deployment. \SystemName builds on conformalized quantile regression (CQR) and utilizes a dedicated loss function to train a quantile function that generates a prediction confidence interval for the single-point output of the DNN model. It advances standard CQR by  dynamically adjusting the prediction interval based on individual locations or sensor nodes of the test samples, leading to a more accurate PI with valid coverage.

We evaluate \SystemName by applying it to six representative DNN architectures for traffic forecasting and compare it against five uncertainty quantification methods. Experimental results show that \SystemName has good generalization ability, delivering better performance than the competing methods across the evaluated DNN models. This outcome underscores the significance of \SystemName as one of the first attempts to develop a generic framework for modeling uncertainties in traffic forecasting.

We will publish the code and data of \SystemName upon acceptance. We hope that \SystemName will enable more research into robust traffic forecasting by providing a means to quantify uncertainty in DNN-based models.

\bibliographystyle{IEEEtran}
\bibliography{reference}
\ifCLASSOPTIONcaptionsoff
  \newpage
\fi
\end{document}